\documentclass{article}
\usepackage{amsmath,graphicx,mlspconf}
\usepackage{cite}
\usepackage{amssymb,amsfonts}
\usepackage{algorithmic}
\usepackage{enumitem}

\usepackage{graphicx}
\usepackage{textcomp}
\usepackage{xcolor}
\usepackage[export]{adjustbox}
\usepackage{booktabs}
\usepackage[ruled,vlined]{algorithm2e}
\usepackage{colortbl}  
\usepackage{multirow}  
\usepackage{hyperref}  
\usepackage{interval}  
\usepackage{bm}
\usepackage{tabularx}  
\newcolumntype{P}[1]{>{\centering\arraybackslash}p{#1}}
\newcolumntype{Y}{>{\centering\arraybackslash}X}  

\makeatletter
\DeclareRobustCommand\onedot{\futurelet\@let@token\@onedot}
\def\@onedot{\ifx\@let@token.\else.\null\fi\xspace}

\def\ie{\emph{i.e}\onedot}

\makeatother

\DeclareMathOperator*{\Motimes}{\text{\raisebox{0.25ex}{\scalebox{0.75}{$\bigotimes$}}}}
\DeclareMathOperator*{\Modot}{\text{\raisebox{0.25ex}{\scalebox{0.75}{$\bigodot$}}}}
\newcommand{\floor}[1]{\lfloor #1 \rfloor}

\copyrightnotice{978-1-6654-8547-0/22//\$31.00 {\copyright}2022 IEEE}

\toappear{2022 IEEE International Workshop on Machine Learning for Signal Processing, Agu.\ 22--25, 2022, Xi'an, China}

\title{C3-SL: Circular Convolution-Based Batch-Wise Compression for Communication-Efficient Split Learning}
\name{Anonymous\thanks{Anonymous.}}
\address{Anonymous}

\name{%
    Cheng-Yen Hsieh$^{\star}$%
    \qquad Yu-Chuan Chuang$^{\dagger}$%
    \qquad An-Yeu (Andy) Wu$^{\star \dagger}$,~\textit{Fellow,~IEEE}%
}
\address{%
    $^{\star}$ Department of Electrical Engineering, National Taiwan University, Taipei, Taiwan \\%
    $^{\dagger}$ Graduate Institute of Electronics Engineering, National Taiwan University, Taipei, Taiwan%
}

\begin{document}

\maketitle

\begin{abstract}
Most existing studies improve the efficiency of Split learning (SL) by compressing the transmitted features. However, most works focus on dimension-wise compression that transforms high-dimensional features into a low-dimensional space. In this paper, we propose \underline{c}ircular \underline{c}onvolution-based batch-wise \underline{c}ompression for SL (C3-SL) to compress multiple features into one single feature. To avoid information loss while merging multiple features, we exploit the quasi-orthogonality of features in high-dimensional space with circular convolution and superposition. To the best of our knowledge, we are the first to explore the potential of batch-wise compression under the SL scenario. Based on the simulation results on CIFAR-10 and CIFAR-100, our method achieves a 16x compression ratio with negligible accuracy drops compared with the vanilla SL. Moreover, C3-SL significantly reduces 1152x memory and 2.25x computation overhead compared to the state-of-the-art dimension-wise compression method.
\end{abstract}


\section{Introduction}
Over the last several years, deep neural networks (DNNs) have been successfully applied in various applications, including computer vision, natural language processing, disease diagnosis. Unfortunately, training entire DNNs requires tremendous computation and memory resources, which hinders the deployment of DNNs on resource-limited edge devices. Therefore, the most common approach to train DNNs is to transmit raw data acquired in edge devices to be processed at cloud servers (cloud-based learning). However, this approach introduces a vast amount of communication overhead and easily intrudes on user’s data privacy during data transmission.

To address this issue, split learning (SL) \cite{vepakomma2018split} is proposed to efficiently train DNNs and protect user’s data privacy by cloud-edge collaborative learning. Generally, SL divides a DNN into two parts between the edge device and the cloud server. The edge device, which owns the user’s data, trains the front part and transmits activations of the last layer (cut layer) to the cloud server. Then, the cloud server trains the remaining part and transmits the gradients of the cut layer back to the edge device to complete one training epoch. In this way, SL can achieve a good balance between on-device computation and communication costs as well as mitigate the privacy concern. 

\begin{figure}[t]
\begin{center}

\includegraphics[width=1\linewidth]{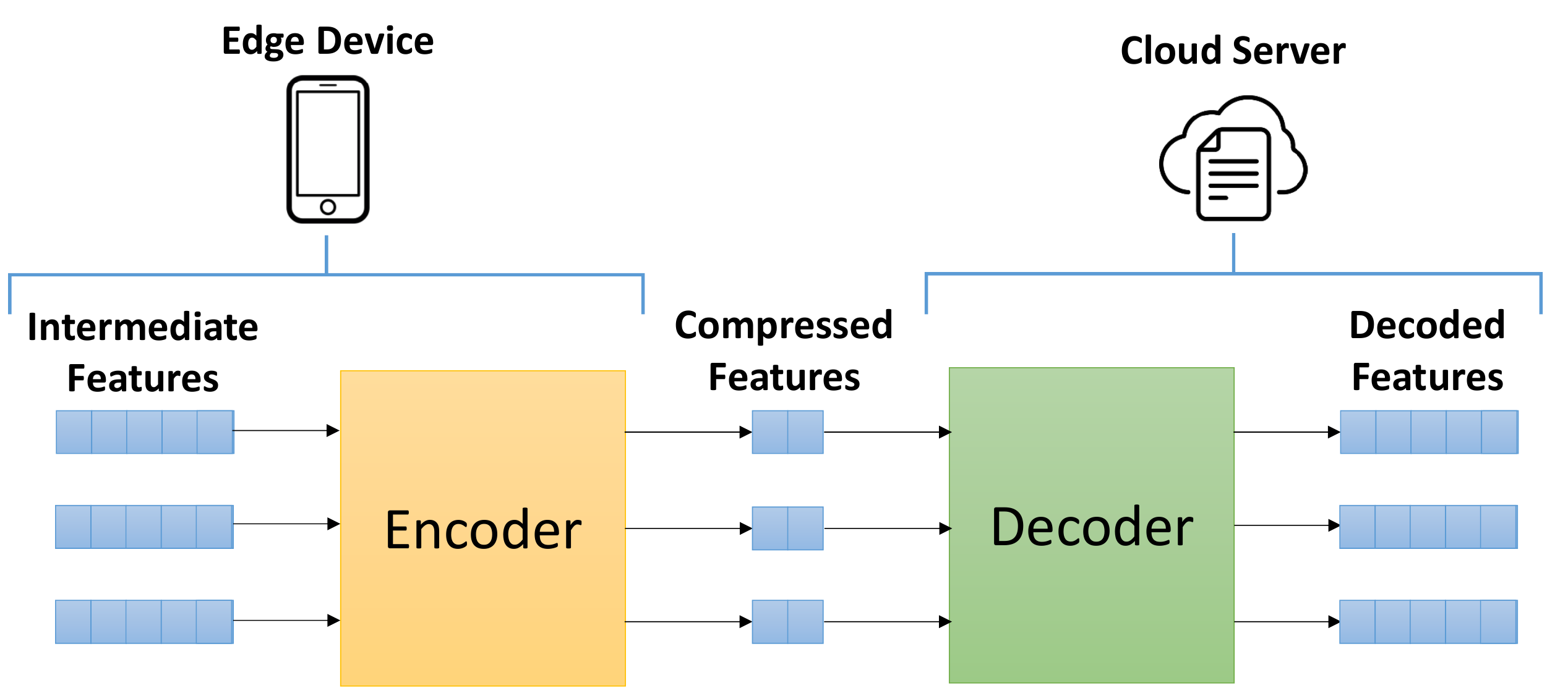}
   \\
   \small (a) 
   \\
\rule{0.2cm}{0.8pt} \rule{0.2cm}{0.8pt} \rule{0.2cm}{0.8pt} \rule{0.2cm}{0.8pt} \rule{0.2cm}{0.8pt} \rule{0.2cm}{0.8pt} \rule{0.2cm}{0.8pt} \rule{0.2cm}{0.8pt} \rule{0.2cm}{0.8pt} \rule{0.2cm}{0.8pt} \rule{0.2cm}{0.8pt} \rule{0.2cm}{0.8pt} \rule{0.2cm}{0.8pt} \rule{0.2cm}{0.8pt} \rule{0.2cm}{0.8pt} \rule{0.2cm}{0.8pt} \rule{0.2cm}{0.8pt} \rule{0.2cm}{0.8pt} \rule{0.2cm}{0.8pt} \rule{0.2cm}{0.8pt} \rule{0.2cm}{0.8pt} \rule{0.2cm}{0.8pt} \rule{0.2cm}{0.8pt} \rule{0.2cm}{0.8pt}
\rule{0.2cm}{0.8pt} \rule{0.2cm}{0.8pt} \rule{0.2cm}{0.8pt} 
\\
\includegraphics[width=1\linewidth]{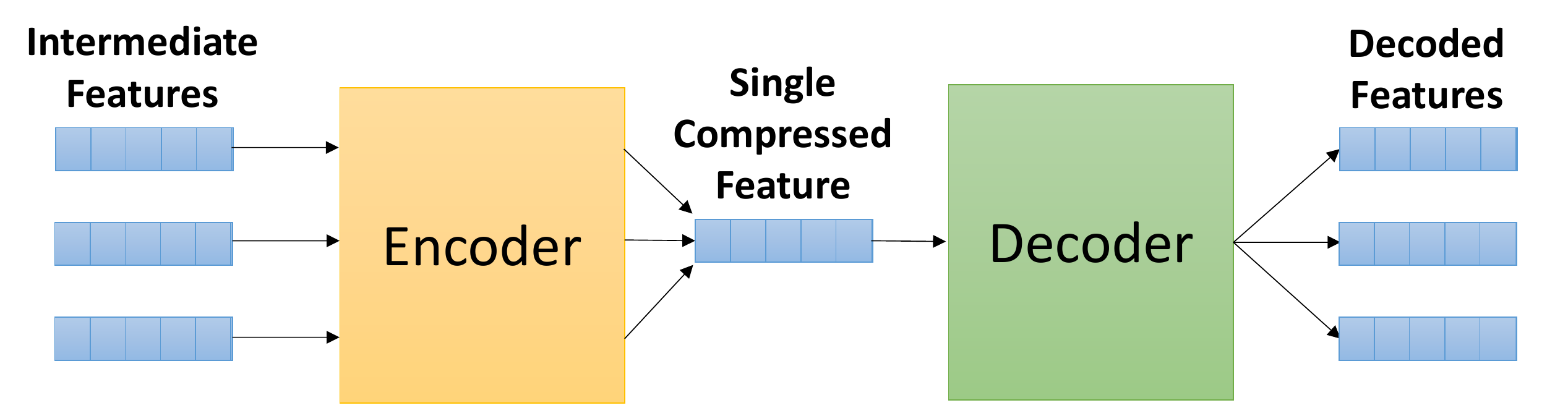}
\\ 
\small (b) 
\end{center}
\caption{(a) Dimension-wise compression vs. (b) the proposed batch-wise compression for split learning.}
\label{fig:DWBW_Compre}
\end{figure}


\begin{figure*}[ht]

    \begin{center}
    \includegraphics[width=1\linewidth]{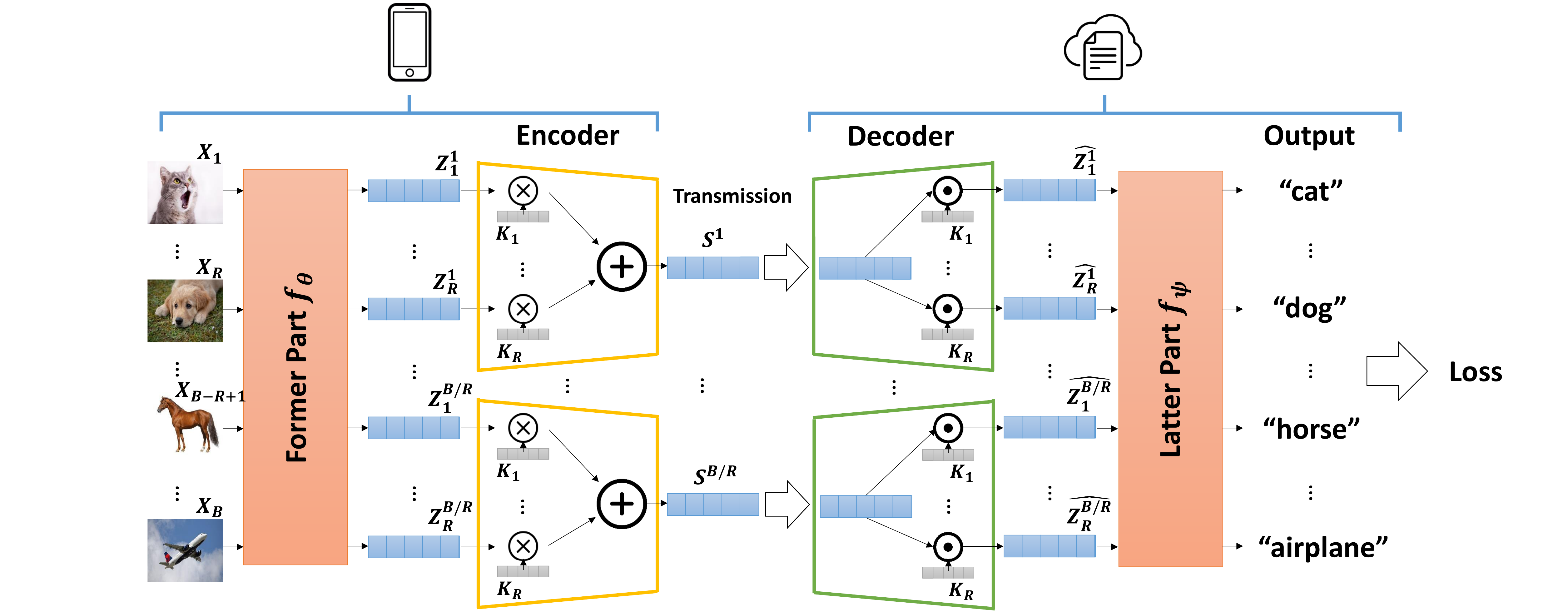}
    \end{center}
\caption{Proposed Framework of C3-SL.}
\label{fig:framework}
\end{figure*}

To further improve the efficiency of SL, several works \cite{matsubara2022supervised,choi2020back,li2018jalad,eshratifar2019bottlenet,chen2019toward,singh2020end,ko2018edge,shao2020bottlenet++} compress the feature dimensions of activations and gradients of the cut layer before transmission to reduce the communication costs. \cite{matsubara2022supervised} utilized knowledge distillation to compress the intermediate features, but the demand of training a large teacher model increases the computation burden. BaF \cite{choi2020back} compressed a selected subset of features and restored the full features by a trainable network. However, raw data must be uploaded to the cloud server to train the restored network, which induces privacy issues and further increases communication costs. On the other hand, J. Shao and J. Zhan~\cite{shao2020bottlenet++} proposed an end-to-end trainable architecture with a DNN-based codec, called BottleNet++, for feature compression, as shown in Fig.~\ref{fig:DWBW_Compre}(a).

It can be shown that most prior works focus on dimension-wise compression while achieving no accuracy degradation. To the best of our knowledge, no work investigates the possibility of batch-wise compression that compresses a batch of features into one feature before transmission, as shown in Fig.~\ref{fig:DWBW_Compre}(b). In this paper, we first propose a circular convolution-based batch-wise compression for SL, called C3-SL, to efficiently reduce communication costs. C3-SL exploits the quasi-orthogonality of features in high-dimensional space and compresses multiple features into a single feature by circular convolution. The main contributions of our work are described as follows:
\begin{itemize}[leftmargin=0.5cm]
    \item [1)] C3-SL is the first framework to exploit batch-wise compression on SL. With circular convolution, C3-SL efficiently and effectively compresses a batch of features and gradients. The proposed C3-SL reduces 16x communication costs with negligible accuracy drops on both CIFAR-10 and CIFAR-100.
    
    \item [2)] Compared with BottleNet++ based on dimension-wise compression, C3-SL reduces the computational and memory overhead by 1152x and 2.25x, respectively.
\end{itemize}

The rest of this paper is organized as follows. Section~\ref{sec:background} reviews the background of split learning and the related work. In Section~\ref{sec:method}, we introduce our proposed circular convolution-based batch-wise compression for SL. The simulation results are discussed in Section~\ref{sec:simulation}. Finally, we conclude our work in Section~\ref{sec:conclusion}.

\section{Background}
\label{sec:background}

\subsection{Split Learning}

The training of split learning consists of two main parts: 1) the splitting step and 2) the training step. The splitting step involves dividing a DNN into two separate parts, where we deploy the former part of the neural network $f_\theta$ in an edge device, and the latter part $f_\psi$ in a cloud server. Both sides initialize the network randomly and proceed to the training step. 
The training step begins from the forward propagation of data through the former part of the neural network $f_\theta$. The output features obtained from $f_\theta$ and the corresponding labels are then transmitted over to the cloud server. The cloud will continue the forward propagation by processing the received features through the latter part of the network $f_\psi$. In the backpropagation process, the cloud server computes the gradients with the transmitted labels and propagates the gradients backward. The gradients generated at the first layer of $f_\psi$ is then transmitted back to the edge device to complete the backpropagation. All these steps will be repeated until completing the training process.

\subsection{Dimension-Wise Compression}
%
Many works~\cite{matsubara2022supervised,choi2020back,li2018jalad,eshratifar2019bottlenet,chen2019toward,singh2020end,ko2018edge,shao2020bottlenet++} have been conducted to compress the dimensions of feature representations at the split layer to reduce the communication costs and latency for feature transmission. Z. Chen \textit{et al.}~\cite{chen2019toward} describes conventional lossy and lossless codecs for deep feature compression. The approach in \cite{ko2018edge} applies JPEG and Huffman Coding to compress the intermediate feature activations. The method in \cite{li2018jalad} quantizes floating-point data and encodes them with Huffman coding. S. Singh \textit{et al.}~\cite{singh2020end} jointly optimizes the network for compressibility and the task objective to increase the compressibility of intermediate features. However, these works~\cite{chen2019toward,ko2018edge,li2018jalad,singh2020end} only adopt dimension-wise feature compression techniques at the inference stage, which causes serious communication burdens at the training stages of SL.

Recently, several methods~\cite{matsubara2022supervised,choi2020back,eshratifar2019bottlenet,shao2020bottlenet++} leverage feature compression techniques to further improve the efficiency of training for SL. Nevertheless, the works in~\cite{matsubara2022supervised,choi2020back} still consume heavy computation or communication costs due to the utilization of extra teacher models or the demands to transmit raw data. Considering the constraints, we select BottleNet++~\cite{shao2020bottlenet++}, detailed in the Section~\ref{sec:BottleNet}, as the main comparison target in this paper. Different from the above works that adopt dimension-wise compression, our method aims to develop a new compression paradigm, \ie, \emph{batch-wise compression}, by compressing multiple features into a single feature vector.

\subsection{BottleNet++}
\label{sec:BottleNet}
BottleNet++~\cite{shao2020bottlenet++} trained the encoder and decoder in an end-to-end manner for feature compression. The encoder consists of a convolutional layer, a batch normalization layer, and a Sigmoid function. In the training stages, the encoder applies lossy compression to reduce the dimension of the intermediate features obtained from $f_\theta$. Specifically, let $(C,W,H)$ denotes the (channel, width, height) of intermediate features. The channel size of features is compressed to $C^{\prime}$ by adopting $C^{\prime}$ filters in the convolutional layer. To compress feature width from $W$ to $W^{\prime}$ and feature height from $H$ to $H^{\prime}$, the stride of convolutional operation is set to 
$(\floor{W/W^{\prime}},\floor{H/H^{\prime}})$. 
After transmitting the compressed features to the cloud server, the decoder restores the compressed features to the original dimension. The decoder consists of a deconvolutional layer, a batch normalization layer, and a ReLU activation function. 
The deconvolutional layer adopts $C$ filters to recover $C^{\prime}$ channels back to $C$ channels. $W$ and $H$ can be restored by adopting
$(\floor{W/W^{\prime}},\floor{H/H^{\prime}})$ stride in convolutional operation. The restored features are then sent to $f_\psi$ to complete the forward propagation.

\section{Proposed Circular Convolution-Based Batch-Wise Compression for SL}

\label{sec:method}

The proposed C3-SL framework is shown in Fig.~\ref{fig:framework}. Let $B$ denotes the batch size, and $\boldsymbol{\mathit{X}} = \{X_1,X_2,\dots,X_B\}$ denotes a batch of input data. Intermediate features $\boldsymbol{\mathit{Z}}\in \mathbb{R}^{B\times D}$ is obtained by $\boldsymbol{\mathit{Z}}=f_\theta(\boldsymbol{\mathit{X}})$, where $D=C\times H\times W$ denotes the number of dimensions of the intermediate features. The intermediate features $\boldsymbol{\mathit{Z}}$ are equally divided into different
groups, where each group contains $R$ features. Each group is transferred to an encoder to be compressed into one single feature. Consequently, we use $R$ to denote the compression ratio. After receiving the compressed features, the decoder in the cloud server decodes each of the compressed features back to original $R$ features. All the restored features will then be concatenated and passed through the remaining layers $f_\psi$ to complete the forward pass. The pseudocode of C3-SL is shown as Algorithm~\ref{alg:Algorithm1}. The details of the encoder and the decoder will be elaborated in Section~\ref{sec3-1:Encoder} and Section~\ref{sec3-2:Decoder}.

\definecolor{codegreen}{rgb}{0.,0.5,0.1}

\begin{algorithm}[t]
\caption{Proposed C3-SL}
\label{alg:Algorithm1}
\SetAlFnt{\small}
\newcommand\mycommfont[1]{\small\textit{\textcolor{codegreen}{#1}}}
\SetAlgoLined
\SetCommentSty{mycommfont}
\SetKwComment{Comment}{\# }{}

\Comment*[h]{
$R$: Batch-wise Compression Ratio (Number of Keys)}
\BlankLine
\Comment*[h]{$B$: Batch Size}
\BlankLine
\Comment*[h]{$f_\theta$: The former part of network in edge device}
\BlankLine
\Comment*[h]{$f_\psi$: The latter part of network in cloud server}
\BlankLine
\Comment*[h]{$D$: Number of Dimensions in the feature}
\BlankLine

$\boldsymbol{\mathit{K}} =$ Generate\_Key($R, D$)  \Comment*[h]{Generate number of $R$ Keys}
\BlankLine
\BlankLine

\While{Training}{
    \For(\Comment*[h]{load a batch of $B$ samples}){$\boldsymbol{\mathit{X}}, \boldsymbol{\mathit{Y}}$ in loader}{ 
        $\boldsymbol{\mathit{Z}}$ = $f_\theta(\boldsymbol{\mathit{X}})$  \Comment*[h]{Compute intermediate features $\boldsymbol{\mathit{Z}}$}\\
        $\boldsymbol{\mathit{Z}}$ = $\boldsymbol{\mathit{Z}}$.flatten(1) \\
        Divide\_Group($\boldsymbol{\mathit{Z}}$)  \Comment*[h]{Divide $\boldsymbol{\mathit{Z}}$ into $B/R$ groups} \\
        \BlankLine
        
        \Comment*[h]{Compress each $Z^g$ $(R,D)$ into single feature} \\
        \For{$g=1,\dots,B/R$}
        {
            $S^g$ = Encode$(Z^g, \boldsymbol{\mathit{K}})$ \\
            \BlankLine
            
            \Comment*[h]{====== Send to Cloud Server ======} \\
            $\widehat{Z_i^g}$ = Decode $(S^g, K_i)$ \textbf{for} $i=1,\dots,R$ \\
            \BlankLine
        }
        \BlankLine
        y\_pred = $f_\psi(\widehat{\boldsymbol{\mathit{Z}}})$  \Comment*[h]{$\widehat{\boldsymbol{\mathit{Z}}}$:concatenated features} \\
        \BlankLine
        \Comment*[h]{Cross Entropy (CE) loss} \\
        loss = CE$(\boldsymbol{\mathit{Y}}$, y\_pred$)$ \\
        \BlankLine
        \Comment*[h]{Optimization} \\
        loss.backward() \\
        optimizer.step()

    }
}
\end{algorithm}

\definecolor{codegreen}{rgb}{0.,0.5,0.1}

\begin{table*}[t]
\begin{center}
\caption{Overall comparison on CIFAR-10/CIFAR-100 under different Compression Ratio $R$.}
\label{table:acc}
\resizebox{1.0\linewidth}{0.85\height}{
\begin{tabularx}{1.25\textwidth}{lYYllYll}
\toprule
  & & \multicolumn{3}{c}{\shortstack{VGG-16\\on CIFAR-10}} & \multicolumn{3}{c}{\shortstack{Resnet-50\\on CIFAR-100}}  \\
\cmidrule(lr){3-5} \cmidrule(lr){6-8}
Method & $R$ & \shortstack[l]{Classification\\Accuracy (\%)} & \shortstack[l]{Number of\\Parameters (x10\textsuperscript{3})} & \shortstack[l]{FLOPs\\(x10\textsuperscript{9})} & \shortstack[l]{Classification\\Accuracy (\%)} & \shortstack[l]{Number of\\Parameters (x10\textsuperscript{3})} & \shortstack[l]{FLOPs\\(x10\textsuperscript{9})} \\
\midrule
Vanilla SL (w/o compression)~\cite{vepakomma2018split} & - & 89.9 & - & - & 63.1 & - & - \\ 
\midrule
\multirow{4}{*}{BottleNet++~\cite{shao2020bottlenet++}} & 2 & 90.5 & 2,360.0 & 1.21 & 63.6 & 9,438.7 & 4.83 \\
 & 4 & 90.4 & 2,098.2 & 0.67 & 62.9 & 8,390.7 & 2.68 \\
  & 8 & 89.8 & 1,049.3 & \textbf{0.34} & 62.6 & 4,195.8 & \textbf{1.34} \\
  & 16 & 89.6 & 524.9 & \textbf{0.17} & 62.5 & 2,098.4 & \textbf{0.67} \\
\midrule
\multirow{4}{*}{\textbf{Proposed C3-SL}} & 2 & 90.3 & \textbf{4.1~\textcolor{codegreen}{($576\times$)}} & \textbf{0.54~\textcolor{codegreen}{($2.24\times$)}} & 63.4 & \textbf{8.2~\textcolor{codegreen}{($1152\times$)}} & \textbf{2.15~\textcolor{codegreen}{($2.25\times$)}} \\
 & 4 & 90.0 & \textbf{8.2~\textcolor{codegreen}{($256\times$)}} & \textbf{0.54~\textcolor{codegreen}{($1.24\times$)}} & 63.3 & \textbf{16.4~\textcolor{codegreen}{($512\times$)}} & \textbf{2.15~\textcolor{codegreen}{($1.25\times$)}} \\
  & 8 & 89.9 & \textbf{16.4~\textcolor{codegreen}{($64\times$)}} & 0.54 & 62.8 & \textbf{32.8~\textcolor{codegreen}{($128\times$)}} & 2.15 \\
  & 16 & 89.6 & \textbf{32.8~\textcolor{codegreen}{($16\times$)}} & 0.54 & 62.3 & \textbf{65.5~~~\textcolor{codegreen}{($32\times$)}} & 2.15 \\
\bottomrule
\end{tabularx}}
\end{center}
\end{table*}

\subsection{Encoder: Circular Convolution}
\label{sec3-1:Encoder}

The encoder of C3-SL utilizes circular convolution~\cite{plate1995holographic} and superposition in the hyperdimensional space to compress intermediate features
$\boldsymbol{\mathit{Z}}^g = \{Z_1^g,Z_2^g,\dots,Z_R^g\}$
in group $g$ into one single feature, where $g \in \interval{1}{\frac{B}{R}}$ and
$\boldsymbol{\mathit{Z}}=\{\boldsymbol{\mathit{Z}}^1,\dots,\boldsymbol{\mathit{Z}}^{\frac{B}{R}}\}$.
First, $R$ keys $\boldsymbol{\mathit{K}} = \{K_1,K_2,\dots,K_R\} \in \mathbb{R}^{R\times D}$
are generated, where each key with $D$ dimensions is sampled from a normal distribution $\mathcal{N}(0, \frac{1}{D})$ and normalized to unit norm. Circular convolution is utilized to bind a feature $Z_i^g$ with $K_i$ to generate a binded feature $V_i^g\in \mathbb{R}^D$, which is computed as

\begin{equation}
    \label{eq:en_1}
    V_i^g = K_i \Motimes Z_i^g
\end{equation}
where $\Motimes$ denotes circular convolution. The binded features $\{V_1^g,V_2^g,\dots,V_R^g\}$ will be quasi-orthogonal to each other even if they are close in their original feature space~\cite{kanerva2009hyperdimensional}. This concept allows the superposition of multiple binded features into one single feature. The compressed feature $S^g \in \mathbb{R}^D$ is then obtained by

\begin{equation}
\label{eq:en_2}
    S^g = \sum_{i=1}^{R}{V_i^g}
\end{equation}

Since all the operations in the encoder are continuous, the gradients can be automatically calculated during the backpropagation. However, C3-SL does not compute the gradients for keys or update the weights of keys. In contrast to the CNN encoder in BottleNet++, whose weights need to be optimized, C3-SL dramatically reduces computation burden by utilizing this property.

\subsection{Decoder: Circular Correlation}
\label{sec3-2:Decoder}

The circular correlation~\cite{plate1995holographic} is the approximate inverse of circular convolution to restore the compressed features. In C3-SL, the decoder uses the circular correlation to retrieve the original features back from the compressed features. After the compressed features from each group are transmitted to the decoder in the cloud server, we obtain the decoded feature $\widehat{Z_i^g}$ by performing circular correlation between $S^g$ and the corresponding key $K_i$:
\begin{equation}
    \widehat{Z_i^g} = K_i \Modot S^g
\end{equation}
where $\Modot$ is the circular correlation. All the retrieved features $\widehat{Z_i^g}$ are concatenated and transferred into the latter part $f_\psi$ to complete the forward pass. 

In the decoding process, the retrieval is noisy due to two sources of error, 1) the error from unbinding itself and 2) the terms originating from other binded features:
\begin{equation}
    \widehat{Z_i^g} = K_i \Modot (K_i \Motimes Z_i^g) + \sum_{j\ne i}{K_i \Modot (K_j \Motimes Z_j^g)}
\end{equation}

Despite the lossy nature of this method, we will demonstrate that the performance drop is negligible in Section~\ref{sec4:setup}.


\section{Simulation Results}
\label{sec:simulation}

\subsection{Experimental Setup}
\label{sec4:setup}

In our experiments, we use VGG-16~\cite{simonyan2014very} trained on CIFAR-10~\cite{krizhevsky2009learning} and Resnet-50~\cite{he2016deep} trained on CIFAR-100~\cite{krizhevsky2009learning}, 
to evaluate the effectiveness of C3-SL on small and large-scale models, as well as datasets. Both datasets consist of 50,000 training images and 10,000 test images. We split ResNet-50 at the output of the third residual block and VGG-16 at the output of the 4-th max-pooling layer. All the networks are trained for 100 epochs with cross-entropy loss and Adam optimizer. We select the learning rate as 0.0001. The batch size $B$ is set as 64. To reproduce BottleNet++, we follow the same structures used in~\cite{shao2020bottlenet++}, where the kernel size $k$ is set to $2\times2$ and the stride is set to $(2,2)$ for both the encoder and decoder. Here, we remove the channel condition layers in~\cite{shao2020bottlenet++} to better exploring the relationships between performance and compression ratio. We evaluate the models with all the 10,000 test images on CIFAR-10~\cite{krizhevsky2009learning} or CIFAR-100~\cite{krizhevsky2009learning}. All the results are averaged over 5 independent trials (\ie, 5 independent models).

\subsection{Analysis of Performance}

We compare the accuracy of C3-SL with that of BottleNet++ under different compression ratios $R$. As shown in Table~\ref{table:acc}, C3-SL achieves 89.9\% accuracy with VGG-16~\cite{simonyan2014very} trained on CIFAR-10~\cite{krizhevsky2009learning} and preserves less than 0.1\% performance drop under an $8\times$ compression ratio compared to vanilla SL, where we train the networks without any compression. The performance of C3-SL slightly degrades under $R=16$, but still converges within a 0.3\% performance drop. To better examine the impact of compression ratio on performance, we further validate the proposed method with a larger architecture Resnet-50~\cite{he2016deep} trained on the CIFAR-100 dataset~\cite{krizhevsky2009learning}. Compared with BottleNet++, C3-SL obtains competitive or even better performance under each compression ratio while achieving low memory and computational overhead, as illustrated in the next section.

\subsection{Analysis of Memory and Computational Overhead}

To demonstrate the efficiency of C3-SL, we compare the memory and computation overhead between C3-SL and BottleNet++. In TABLE~\ref{table:overhead}, we provide formulas to compute the required number of parameters and FLOPs of different frameworks. Specifically, the number of parameters from generating $R$ keys is $R\times D$ in C3-SL, where each key contains $D$ dimensions. The circular convolution and circular correlation both consume $D^2$ FLOPs. Consequently, $2BD^2$ FLOPs of C3-SL are required for each batch of training.

One of the main advantages of C3-SL is that it dramatically reduces the memory overhead. As shown in TABLE~\ref{table:acc}, compared to BottleNet++, C3-SL dramatically reduces $16\times$ and $32\times$ memory overhead under $R=16$ with VGG-16~\cite{simonyan2014very} on CIFAR-10~\cite{krizhevsky2009learning} and Resnet-50~\cite{he2016deep} on CIFAR-100~\cite{krizhevsky2009learning}, respectively. Moreover, C3-SL achieves $1152\times$ less memory overhead and reduces $2.25\times$ computation burden under $R=2$ on CIFAR-100~\cite{krizhevsky2009learning}. This indicates that under the scenario of limited memory and computation resource, C3-SL is more suitable for edge devices to compress the intermediate features.

\section{conclusion}
\label{sec:conclusion}

In this work, C3-SL exploits circular convolution and superposition to compress multiple features into a single feature before transmission. Compared with the dimension-wise compression BottleNet++, C3-SL significantly reduces $1152\times$ memory overhead and $2.25\times$ computation burden. This advantage indicates that C3-SL is a promising framework for the future trend of SL while achieving high memory and computation efficiency. In future work, we will explore the potential of combining dimension-wise and batch-wise compression to further reduce communication costs.

\begin{table}
\caption{Computation and Memory Overhead of Different Frameworks}
\label{table:overhead}
\centering
\vspace{3mm}
\resizebox{1.0\linewidth}{!}{

    \begin{tabular}{|c|c|c|} 
    \hline
    \textbf{Method} & \begin{tabular}[c]{@{}c@{}}\textbf{Number of }\\\textbf{Parameters}\end{tabular} & \begin{tabular}[c]{@{}c@{}}\textbf{FLOPs} \\\textbf{(Training stages)}\end{tabular}  \\ 
    \hline
    BottleNet++     & $\begin{aligned}\\[-2ex]
    \frac{\left(Ck^2+1\right) 4C}{R} + \\ 
    \left(\frac{4C}{R}k^2+1\right)C \\[1ex]
    \end{aligned}$ & $\begin{aligned}\\[-2ex]\frac{B \left(2Ck^2+1\right) 4CH^{\prime}W^{\prime}}{R} + \\B \left(\frac{8C}{R}k^2+1\right) CHW \\[1ex]\end{aligned}$ \\
    \hline
    \rule{0pt}{12pt}\begin{tabular}[c]{@{}c@{}}\textbf{Proposed}\\\textbf{C3-SL}\end{tabular}  & $R\times D$ & $2BD^2$ \\
    \hline
    \end{tabular}
}

\end{table}


\section{Acknowledgement}
This work was supported by the Ministry of Science and Technology of Taiwan under Grant MOST 111-2218-E-002-018-MBK and 110-2218-E-002-034-MBK.

\bibliographystyle{IEEEbib}
\bibliography{refs}

\end{document}